\pdfoutput=1

\documentclass[11pt]{article}

\usepackage{acl}

\usepackage{times}
\usepackage{latexsym}

\usepackage[T1]{fontenc}

\usepackage[utf8]{inputenc}

\usepackage{microtype}

\usepackage{multirow}
\usepackage[inline]{enumitem}
\usepackage{graphicx}
\usepackage{amssymb}
\usepackage{pifont}
\usepackage{pgfplots}
\usetikzlibrary{patterns}

\pgfplotsset{compat=1.17}

\newcommand{\vp}{VoxPopuli}
\newcommand{\epst}{Europarl-ST}
\newcommand{\origunit}{\textit{orig-unit}}
\newcommand{\ftunit}{\textit{norm-unit}}
\newcommand{\cmark}{\ding{51}}
\newcommand{\xmark}{\ding{55}}

\title{Textless Speech-to-Speech Translation on Real Data}

\author{
Ann Lee, Hongyu Gong, Paul-Ambroise Duquenne, Holger Schwenk, \\
{\bf Peng-Jen Chen, Changhan Wang, Sravya Popuri, Yossi Adi,} \\
{\bf Juan Pino, Jiatao Gu, Wei-Ning Hsu} \\ \\
Meta AI\\
\texttt{\{annl,wnhsu\}@fb.com}
}

\begin{document}
\maketitle
\begin{abstract}
We present a textless speech-to-speech translation (S2ST) system that can translate speech from one language into another language and can be built without the need of any text data.
Different from existing work in the literature, we tackle the challenge in modeling multi-speaker target speech and train the systems with real-world S2ST data.
The key to our approach is a self-supervised unit-based speech normalization technique, which finetunes a pre-trained speech encoder with paired audios from multiple speakers and a single reference speaker to reduce the variations due to accents, while preserving the lexical content.
With only 10 minutes of paired data for speech normalization, we obtain on average 3.2 BLEU gain when training the S2ST model on the \vp~S2ST dataset, compared to a baseline trained on un-normalized speech target.
We also incorporate automatically mined S2ST data and show an additional 2.0 BLEU gain.
To our knowledge, we are the first to establish a textless S2ST technique
that can be trained with real-world data and works for multiple language pairs\footnote{Audio samples are available at \url{https://facebookresearch.github.io/speech_translation/textless_s2st_real_data/index.html}}.

\end{abstract}

\section{Introduction}
Speech-to-speech translation (S2ST) technology can help bridge the communication gap between people speaking different languages. Conventional S2ST systems~\citep{lavie1997janus,nakamura2006atr} usually rely on a cascaded approach by first translating speech into text in the target language, either with automatic speech recognition (ASR) followed by machine tranlsation (MT), or an end-to-end speech-to-text translation (S2T) model~\citep{berard2016listen}, and then applying text-to-speech (TTS) synthesis to generate speech output.

On the other hand, researchers have started exploring direct S2ST~\citep{jia2019direct,jia2021translatotron,tjandra2019speech,zhang2020uwspeech,kano2021transformer,lee2021direct}, which aims at translating speech in the source language to speech in the target language without the need of text generation as an intermediate step.
However, text transcriptions or phoneme annotations of the speech data is often still needed during model training for multitask learning~\citep{jia2019direct,lee2021direct} or for learning a decoder that generates intermediate representations~\citep{jia2021translatotron,kano2021transformer} to facilitate the generation of speech output.

More than 40\% of the languages in the world are without text writing systems\footnote{\url{https://www.ethnologue.com/}}, while very limited work exist to tackle the challenge of training direct S2ST systems without the use of any text data~\citep{tjandra2019speech,zhang2020uwspeech}.
Moreover, due to the lack of S2ST training data, previous work on direct S2ST mainly rely on TTS to generate synthetic target speech for model training.
The recent release of the large-scale S2ST data from \vp~\citep{wang-etal-2021-voxpopuli} has opened up the possibility of conducting S2ST research on real data.
In addition,~\citet{duquenne2021multimodal} have demonstrated the first proof of concept of direct S2S mining without using ASR or MT systems.
The approach may potentially mitigate the data scarcity issue, but the authors had not evaluated the usefulness of such data for S2ST frameworks.

Most recently,~\citet{lee2021direct} have proposed to take advantage of self-supervised discrete representations~\citep{lakhotia2021generative}, or discrete units, learned from unlabeled speech data as the target for building a direct S2ST model.
Experiments conducted with synthetic target speech data have shown significant improvement for translation between unwritten languages.
In this work, we extend the \textit{textless} S2ST setup in~\citep{lee2021direct}, i.e.~training an S2ST system without the use of any text or phoneme data, and conduct experiments on real S2ST datasets, including~\vp~\citep{wang-etal-2021-voxpopuli} and automatically mined S2ST data~\citep{duquenne2021multimodal}.
To tackle the challenge of modeling real target speech where there are multiple speakers with various accents, speaking styles and recording conditions, etc., we propose a speech normalization technique that finetunes a self-supervised pre-trained model for speech with a limited amount of parallel multiple-to-single speaker speech.
Experiments on four language pairs show that when trained with the normalized target speech obtained from a speech normalizer trained with 10-min parallel data, the performance of a textless S2ST model can be improved by 3.2 BLEU points on average compared with a baseline with un-normalized target speech.

The main contributions of this work include:
\begin{itemize}
    \item We propose a speech normalization technique based on self-supervised discrete units that can remove the variations in speech from multiple speakers without changing the lexical content. We apply the technique on the target speech of real S2ST data and verify its effectiveness in the context of textless S2ST.
    \item We empirically demonstrate that with the speech normalization technique, we can further improve a textless S2ST system's performance by augmenting supervised S2ST data with directly mined S2ST data, demonstrating the usefulness of the latter.
    \item To the best of our knowledge, we are the first to establish a textless S2ST technique that can be trained with real-world data, and the technique works for multiple language pairs.
\end{itemize}

\section{Related work}
\paragraph{Direct S2ST}
\citet{jia2019direct,jia2021translatotron} propose a sequence-to-sequence model with a speech encoder and a spectrogram decoder that directly translates speech from one language into another language without generating text translation first. The model can be trained end-to-end, while phoneme data is required in model training.
On the other hand,
\citet{tjandra2019speech,zhang2020uwspeech} build direct S2ST systems for languages without text writing systems by adopting Vector-Quantized Variational Auto-Encoder (VQ-VAE)~\citep{van2017neural} to convert target speech into discrete codes and learn a speech-to-code translation model.
Most recently,~\citet{lee2021direct} propose a direct S2ST system that predicts self-supervised discrete representations of the target speech. The system, when trained without text data, outperforms VQ-VAE-based approach in~\citet{zhang2020uwspeech}.
As a result, in this work, we follow the design in~\citet{lee2021direct} and focus on training direct S2ST systems with real data.

\paragraph{S2ST data}
\vp~\citep{wang-etal-2021-voxpopuli} provides 17.3k hours of S2ST data from European parliament plenary sessions and the simultaneous interpretations for more than 200 language directions, the largest to-date.
There exists few S2ST corpora as the creation process requires transcribing multilingual speech~\citep{tohyama2004ciair,bendazzoli2005approach,zanon-boito-etal-2020-mass} or high-quality ASR models~\citep{wang-etal-2021-voxpopuli}.
On the other hand,~\citet{duquenne2021multimodal} extend distance-based bitext mining~\citep{ccmatrix:2021:acl} to the audio domain by first learning a joint embedding space for text and audio, where sentences with similar meaning are close, independent of the modality or language. The technique was applied to mine for speech-to-speech alignment in LibriVox\footnote{\url{https://librivox.org/api/info}}, creating 1.4k hours of mined S2ST data for six language pairs.
The usefulness of the S2ST datasets is often showcased indirectly through a speech retrieval task~\citep{zanon-boito-etal-2020-mass} or human evaluation of the data quality~\citep{duquenne2021multimodal}, since existing direct S2ST systems are mostly trained with synthetic target speech~\citep{jia2019direct,tjandra2019speech,zhang2020uwspeech,lee2021direct,jia2021translatotron}.
In this work, we develop an S2ST system that can be trained on real target speech to mitigate the discrepancy between the S2ST system and corpus development process.

\paragraph{Speech normalization}
Speech normalization reduces the variation of factors not specified at the input when building TTS systems. One manual approach is to use clean data from a single speaker with minimal non-textual variation~\citep{wang2017tacotron,shen2018natural,ren2019fastspeech,ljspeech17}.
For automatic methods, silence removal with voice activity detection (VAD) is a fundamental approach~\citep{gibiansky2017deep, hayashi2020espnet, wang2021fairseq}.
Speech enhancement can remove the acoustic condition variation when building TTS models with noisy data~\citep{botinhao2016speech,adiga2019speech}.
Speaker normalization through voice conversion, which maps target speech into the same speaker as the source speech in the context of S2ST~\citep{jia2021translatotron}, can be considered as another speech normalization method.
In this work, we propose a novel speech normalization technique based on self-supervised discrete units, which maps speech with diverse variation to units with little non-textual variation.

\section{System}
\label{sec:system}
We follow~\citet{lee2021direct} to use HuBERT~\citep{hsu2021hubert} to discretize target speech and build a sequence-to-sequence speech-to-unit translation (S2UT) model.
We describe the proposed speech normalization method and the S2UT system below.

\begin{figure}[t!]

\begin{minipage}[b]{.9\linewidth}
  \centering
  \centerline{\includegraphics[height=2.5cm]{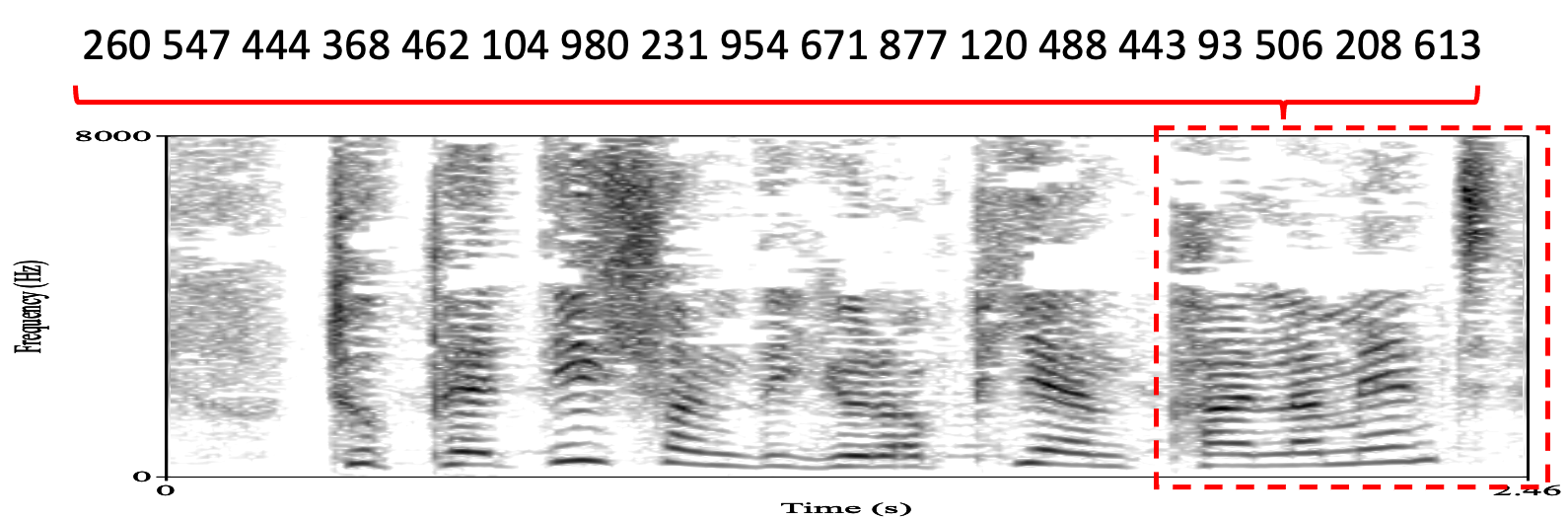}}
  \centerline{(a)}\medskip
\end{minipage}
\hfill
\begin{minipage}[b]{0.9\linewidth}
  \centering
  \centerline{\includegraphics[height=2.5cm]{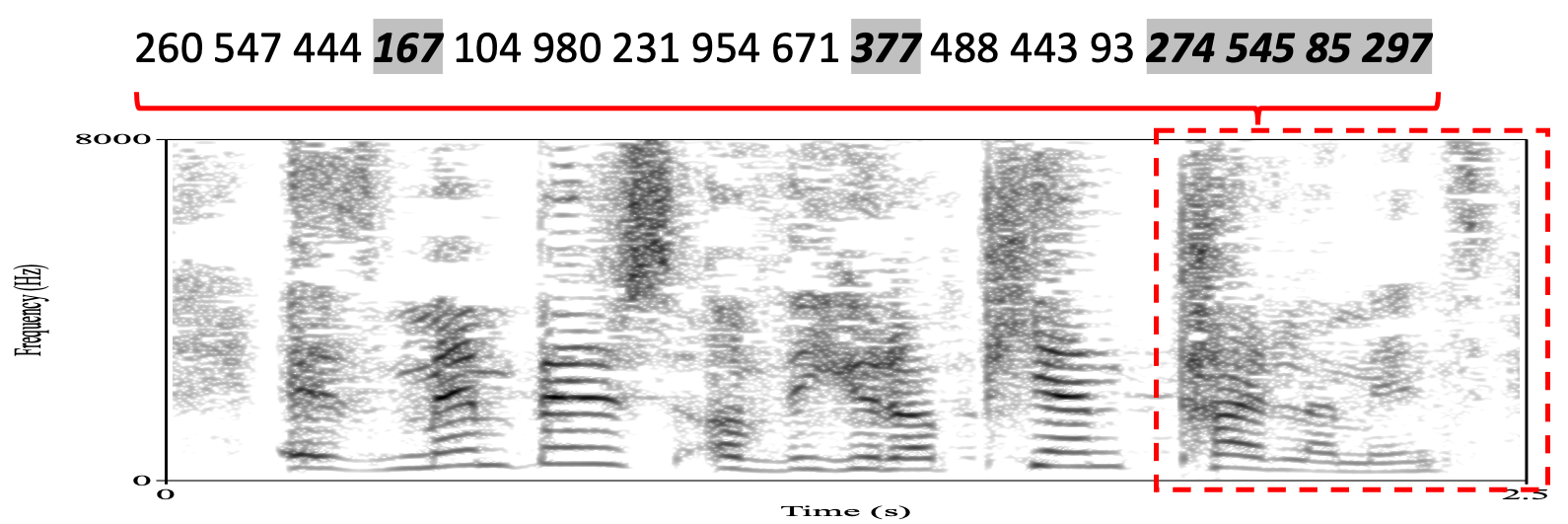}}
  \centerline{(b)}\medskip
\end{minipage}
\hfill
\begin{minipage}[b]{0.9\linewidth}
  \centering
  \centerline{\includegraphics[height=2.5cm]{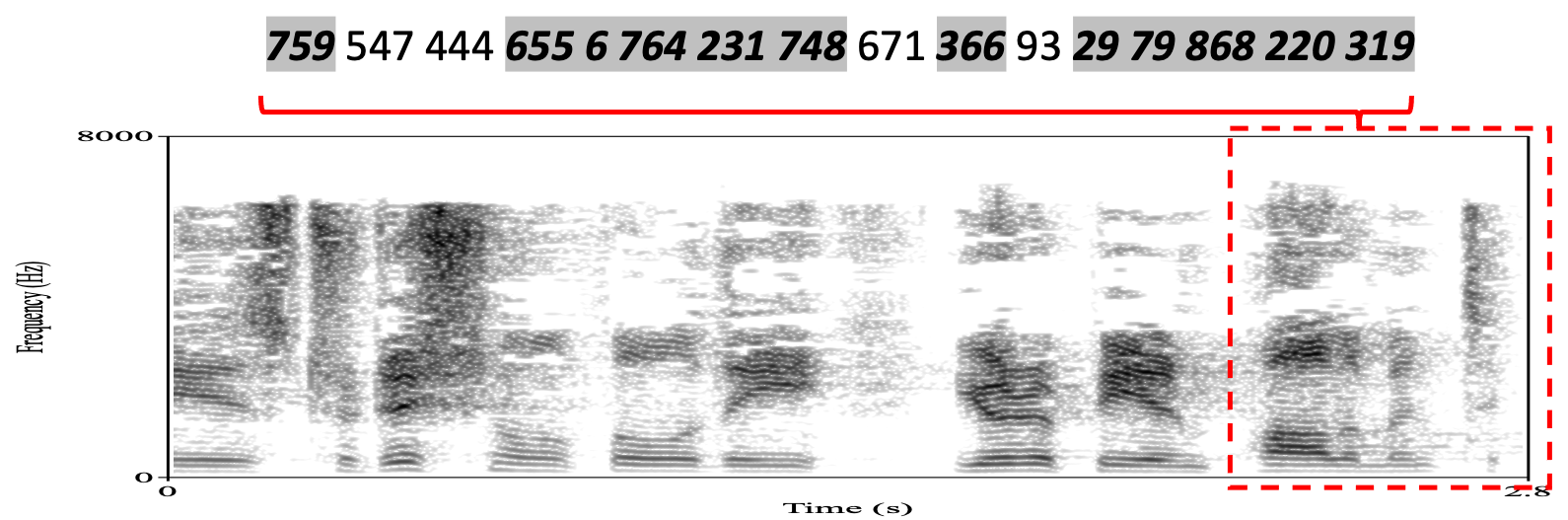}}
  \centerline{(c)}\medskip
\end{minipage}
\caption{Audio samples from one female ((a), (b)) and one male speaker ((c)) from~\vp~\citep{wang-etal-2021-voxpopuli} for the word ``parliament'' and the \textit{reduced} units (consecutive duplicate units removed) encoded by the HuBERT model in Section~\ref{sec:mhubert}.
Differences in the units with respect to (a) are marked in gray.}
\label{fig:unit_samples}
\end{figure}

\subsection{Self-supervised Unit-based Speech Normalization}
\paragraph{HuBERT and discrete units}
Hidden-unit BERT (HuBERT)~\citep{hsu2021hubert} takes an iterative process for self-supervised learning for speech.
In each iteration, K-means clustering is applied on the model's intermediate representations (or the Mel-frequency cepstral coefficient features for the first iteration) to generate discrete labels for computing a BERT-like~\citep{devlin-etal-2019-bert} loss.
After the last iteration, K-means clustering is performed again on the training data, and the learned $K$ cluster centroids are used to transform audio into a sequence of cluster indices as $[z_1, z_2, ..., z_T], z_i \in \{0, 1, ..., K-1\}, \forall 1 \leq i \leq T$, where $T$ is the number of frames.
We refer to these units as \origunit.

\begin{figure}[t!]
    \centering
    \includegraphics[width=0.95\linewidth]{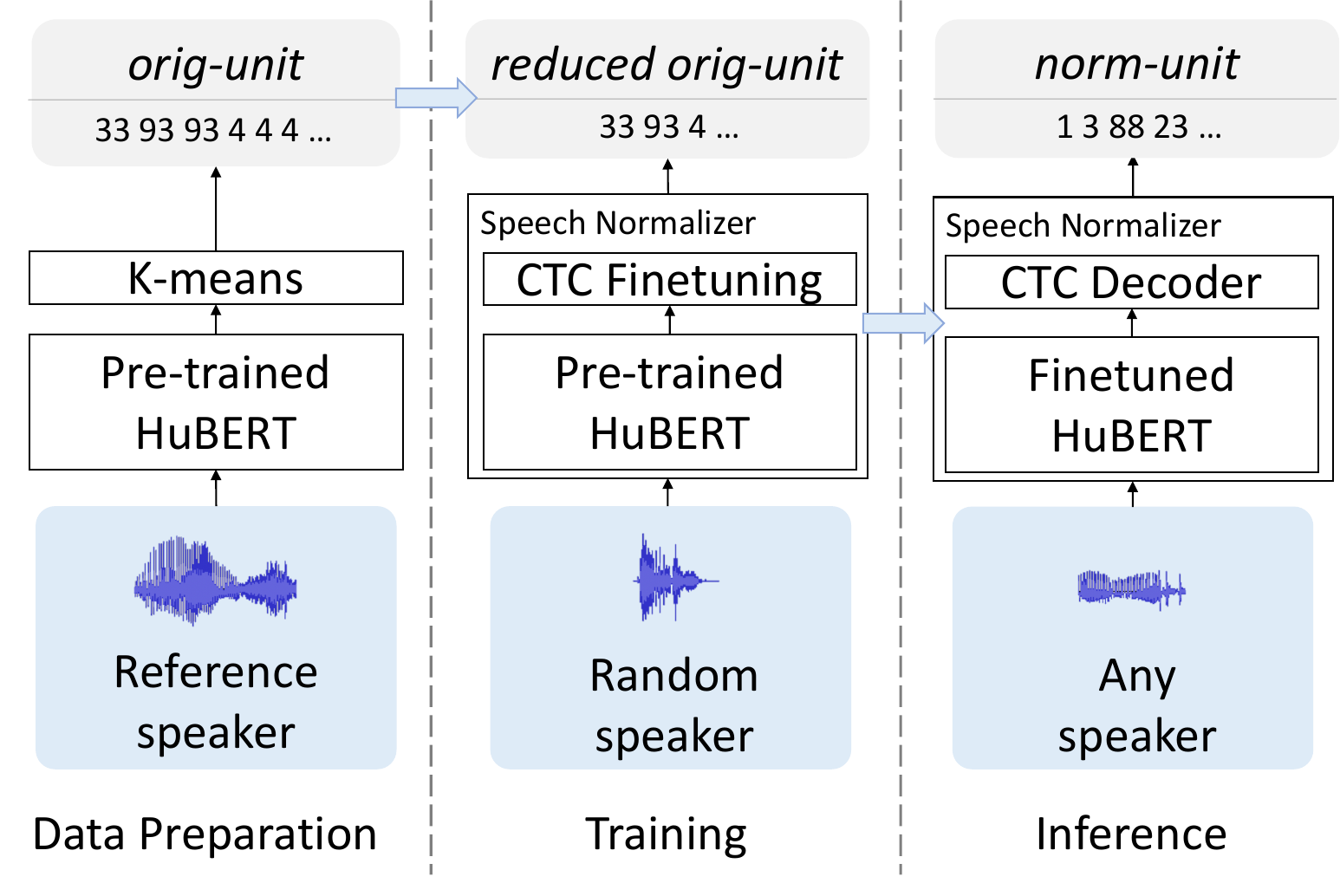}
\caption{Illustration of the self-supervised unit-based speech normalization process.
Left: \origunit~sequences extracted for audios from the reference speaker.
Middle: CTC finetuning with \textit{reduced}~\origunit~from the reference speaker as the target and input audio from different speakers speaking the same content.
Right: For inference, we apply the finetuned speech normalizer and obtain~\ftunit~from CTC decoding.}
\label{fig:speech_norm}
\end{figure}

\begin{figure*}[t!]
    \centering
    \includegraphics[width=0.99\linewidth]{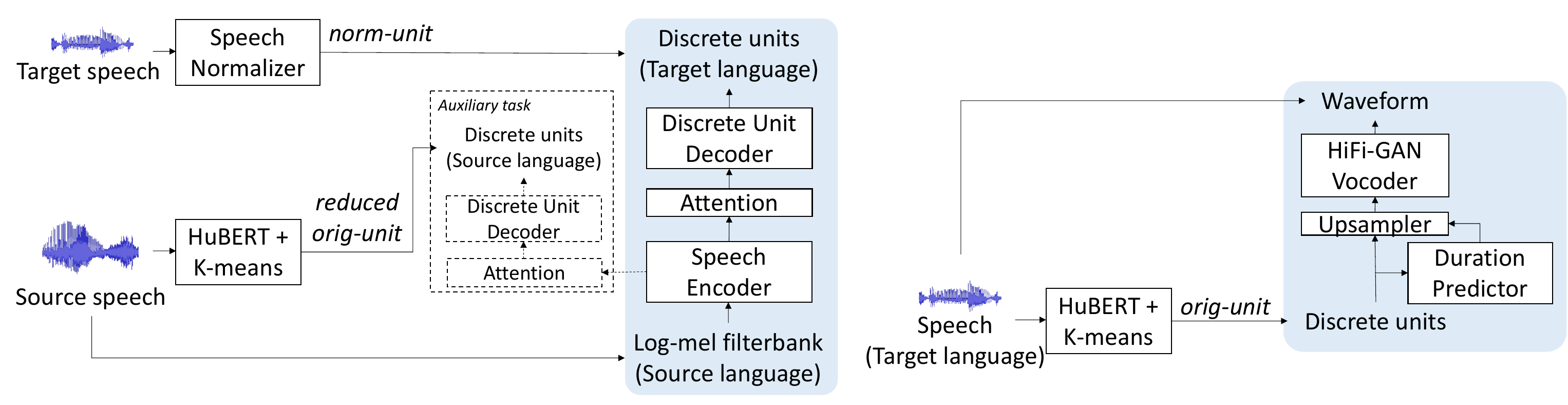}
    \caption{Illustration of the textless S2ST model.
    Left: The speech-to-unit translation (S2UT) model with an auxiliary task.
    Right: The unit-based HiFi-GAN vocoder for unit-to-speech conversion.
    We apply the speech normalizer (Fig.~\ref{fig:speech_norm}) to generate~\ftunit~as the target for S2UT training. The vocoder is  trained with~\origunit~obtained from HuBERT and K-means model. Only the shaded modules are used during inference.
    }
    \label{fig:system}
\end{figure*}

\paragraph{Unit-based speech normalization}
We observe that~\origunit~from audios of different speakers speaking the same content can be quite different due to accent and other residual variations such as silence and recording conditions, while there is less variation in \origunit~from speech from the same speaker (\autoref{fig:unit_samples}).
Following the success of self-supervised pre-training and Connectionist Temporal Classification (CTC) finetuning for ASR~\citep{graves2006connectionist,baevski2019effectiveness}, we propose to build a speech normalizer by performing CTC finetuning with a pre-trained speech encoder using multi-speaker speech as input and discrete units from a reference speaker as target.

\autoref{fig:speech_norm} illustrates the process. 
First, a pair of audios from a random speaker and a reference speaker speaking the same content is required.
Then, we convert the reference speaker speech into \origunit~with the pre-trained HuBERT model followed by K-means clustering.
We further reduce the full \origunit~sequence by removing repeating units~\citep{lakhotia2021generative,lee2021direct,kharitonov2021text,kreuk2021textless}.
The resulting \textit{reduced} \origunit~serves as the target in the CTC finetuning stage with the speech from the random speaker as the input.

The process can be viewed as training an ASR model with the ``pseudo text'', i.e.~units from speech from a single reference speaker.
The resulting speech normalizer is a discrete unit extractor that converts the input speech to units with CTC decoding. We refer to these units as \ftunit.

\subsection{Textless S2ST}
\autoref{fig:system} shows the main components of the system.

\paragraph{Speech encoder}
The speech encoder is built by pre-pending a speech downsampling module to a stack of Transformer blocks~\citep{vaswani2017attention}. The downsampling module consists of two 1D-convolutional layers, each with stride 2 and followed by a gated linear unit activation function, resulting in a downsampling factor of 4~\citep{synnaeve2019end} for the log-mel filterbank input.

\paragraph{Discrete unit decoder}
We train the S2UT system with~\ftunit~as the target.
The unit decoder is a stack of Transformer blocks as in MT~\citep{vaswani2017attention} and is trained with cross-entropy loss with label smoothing. The setup can be viewed as the same as the ``\textit{reduced}'' strategy in~\citet{lee2021direct}, as the speech normalizer is trained on \textit{reduced}~\origunit~sequences.

\paragraph{Auxiliary task}
We follow the unwritten language setup in~\citet{lee2021direct} and incorporate an auto-encoding style auxiliary task to help the model converge during training.
We add a cross-attention module and a Transformer decoder to an intermediate layer of the speech encoder and use \textit{reduced}~\origunit~of the source speech as the target.

\paragraph{Unit-based vocoder}
The unit-to-speech conversion is done with the discrete unit-based HiFi-GAN vocoder~\citep{kong2020hifi} proposed in~\citet{polyak2021speech}, enhanced with a duration prediction module~\citep{ren2020fastspeech}. The vocoder is trained separately from the S2UT model with the combination of the generator-discriminator loss from HiFi-GAN and the mean square error (MSE) of the predicted duration of each unit in logarithmic domain.

\section{Experimental Setup}
\label{sec:exp_setting}
We examine four language pairs: Spanish-English (Es-En), French-English (Fr-En), English-Spanish (En-Es), and English-French (En-Fr).
All experiments are conducted using \texttt{fairseq}~\citep{ott2019fairseq,wang2020fairseqs2t,wang2021fairseqs2}\footnote{Code is available at \url{https://github.com/pytorch/fairseq/blob/main/examples/speech_to_speech/docs/textless_s2st_real_data.md}}.

\subsection{Data}
\label{sec:data}
\paragraph{Multilingual HuBERT (mHuBERT)}
As we focus on modeling target speech in En, Es or Fr, we train a single mHuBERT model (Section~\ref{sec:mhubert}) by combining data from three languages. We use the 100k subset of \vp~unlabeled speech~\citep{wang-etal-2021-voxpopuli}, which contains 4.5k hrs of data for En, Es and Fr, respectively, totaling 13.5k hours.

\begin{table}[t]
\centering
\resizebox{\linewidth}{!}{
\begin{tabular}{c|c|ccc}
 & duration & En & Es & Fr \\
\hline
\hline
\multirow{5}{*}{train} & 10 mins & 89 & 97 & 86 \\
\cline{2-5}
 & \multirow{2}{*}{1 hr} & \multirow{2}{*}{522} & 612 & \multirow{2}{*}{510} \\
 & & & (61\% CV) \\
\cline{2-5}
 & \multirow{2}{*}{10 hrs} & \multirow{2}{*}{5.1k} & 6.7k & 5.9k\\
 & & & (96\% CV) & (56\% CV) \\
\hline
dev & - & 1.2k & 1.5k & 1.5k \\
\end{tabular}}
\caption{\label{tab:accent_ft_data} Number of samples of the data used in training speech normalizers. For Es and Fr, as there is no enough data from \vp~ASR dataset after filtering out the overlap with the S2ST data, we include random samples from the Common Voice 7.0 (CV)~\citep{ardila2020common} dataset (denoted as X\%).}
\end{table}

\paragraph{Speech normalization}
We use multi-speaker speech from the~\vp~ASR dataset~\citep{wang-etal-2021-voxpopuli} and convert text transcriptions to reference units for training the speech normalizer.
The text-to-unit (T2U) conversion is done with a Transformer MT model~\citep{vaswani2017attention} trained on single-speaker TTS data (described later) with characters as input and \textit{reduced}~\origunit~as target.

We build training sets of three different sizes (10-min, 1-hr, 10-hr) for each language (\autoref{tab:accent_ft_data}).
We remove the audios that exist in the \vp~S2ST dataset (described later) and randomly sample from the Common Voice ASR dataset~\citep{ardila2020common} if there is no enough data.
We also randomly sample 1000 audios from Common Voice dev sets and combine with the filtered \vp~ASR dev sets for model development.
Though the reference target is created synthetically, we believe that collecting a maximum of 10-hr speech from a single speaker is reasonable as in TTS data collection~\citep{ljspeech17,park2019css10}.

\begin{table*}[t]
\centering
\resizebox{\linewidth}{!}{
\begin{tabular}{c|cccc|cccc|cccc|cccc}
 & \multicolumn{4}{c|}{Es-En} & \multicolumn{4}{c|}{Fr-En} & \multicolumn{4}{c|}{En-Es} & \multicolumn{4}{c}{En-Fr} \\
\hline
 & \multirow{2}{*}{VP} & \multirow{2}{*}{mined} & \multicolumn{2}{c|}{EP} & \multirow{2}{*}{VP} & \multirow{2}{*}{mined} & \multicolumn{2}{c|}{EP} & \multirow{2}{*}{VP} & \multirow{2}{*}{mined} & \multicolumn{2}{c|}{EP} & \multirow{2}{*}{VP} & \multirow{2}{*}{mined} & \multicolumn{2}{c}{EP} \\
 & & & dev & test & & & dev & test & & & dev & test & & & dev & test \\
\hline
\hline
\# samples & 159k & 314k & 1.9k & 1.8k & 156k & 338k & 1.5k & 1.8k & 126k & 314k & 1.3k & 1.3k &  138k & 338k & 1.3k & 1.2k \\
source (hrs) & 532.1 & 441.7 & 5.4 & 5.1 & 522.9 & 447.1 & 3.7 & 4.7 & 414.7 & 424.7 & 3.0 & 2.9  & 450.6 & 469.5 & 3.0 & 2.8 \\
target (hrs) & 513.1 & 424.7 & 5.6$^\ast$ & - & 507.3 & 469.5 & 3.9$^\ast$ & -  & 424.1 & 441.7 & 3.0$^\ast$  & -  & 456.0 & 447.1 & 3.0$^\ast$ & -   \\
\end{tabular}}
\caption{\label{tab:s2s_data} Statistics of the data used in S2UT model training. We train the models on \vp~(VP)~\citep{wang-etal-2021-voxpopuli} and mined S2ST data~\citep{duquenne2021multimodal} and evaluate on \epst~(EP)~\citep{iranzo2020europarl}. The source speech from plenary sessions before 2013 are removed from VP to avoid overlap with EP, resulting in different amounts of data between X-Y and Y-X language pairs. ($\ast$: speech is created with TTS for tracking dev loss during training.) }
\end{table*}

\paragraph{S2UT}
We use the \vp~S2ST dataset~\citep{wang-etal-2021-voxpopuli} as the supervised S2ST data for model training.
Take Es-En for example. We combine data from Es source speech to En interpretation with Es interpretation to En source speech for training.
We evaluate on the dev set and test set from \epst~\citep{iranzo2020europarl}, as it provides text translation for BLEU score computation and is of the same domain as~\vp.
In addition, we investigate incorporating S2ST data automatically mined from LibriVox~\citep{duquenne2021multimodal}.\footnote{\url{https://github.com/facebookresearch/LASER}}
\autoref{tab:s2s_data} summarizes the statistics of the data for each language pair.

\begin{table}[t!]
\centering
\resizebox{\linewidth}{!}{
\begin{tabular}{c|c|cc}
 & \multirow{2}{*}{dataset} & \multicolumn{2}{c}{duration (hrs)}\\
 & & train & dev \\
\hline
\hline
En & LJSpeech~\citep{ljspeech17} & 22.3 & 0.7 \\
Es & CSS10~\citep{park2019css10} & 20.8 & 0.2 \\
Fr & CSS10~\citep{park2019css10} & 17.7 & 0.2 \\
\end{tabular}}
\caption{\label{tab:tts_data} Duration of the TTS datasets after VAD. }
\end{table}

\paragraph{TTS data}
We train the unit-based HiFi-GAN vocoder using TTS data, pre-processed with VAD to remove silence at both ends of the audio. No text data is required during vocoder training.
In addition, we use the same TTS dataset to train the T2U model for generating reference target units in speech normalizer training and to build the cascaded baselines described in Section~\ref{sec:baseline}.

\subsection{Multilingual HuBERT (mHuBERT)}
\label{sec:mhubert}
We build a single mHuBERT model for all three languages using the combination of 13.5k-hr data without applying any language-dependent weights or sampling, since the amount of data is similar between all three languages.
A single codebook is used for all three languages, and no language information is required during pre-training. The mHuBERT model is pre-trained for three iterations following~\citet{hsu2021hubert,lakhotia2021generative}.
In each iteration, model weights are randomly initialized and optimized for 400k steps.
We find that $K=1000$ with features from the 11-th layer of the third-iteration mHuBERT model work the best for our experiments.

\subsection{Baselines}
\label{sec:baseline}
\paragraph{S2UT with \textit{reduced}~\origunit}
First, we consider a basic setup by training the S2UT system using \textit{reduced} \origunit~extracted from the target multi-speaker speech with mHuBERT~\citep{lee2021direct}.
For the second baseline, we concatenate a d-vector speaker embedding~\citep{variani2014deep} to each frame of the speech encoder output to incorporate target speaker information.
A linear layer is applied to map the concatenated feature vectors to the same dimension as the original encoder output.
The 256-dimensional speaker embedding, which remains fixed during the S2UT model training, is extracted from a speaker verification model pre-trained on VoxCeleb2~\citep{chung2018voxceleb2}. During inference, we use the speaker embedding averaged from all audios from the TTS dataset of the target language.

\paragraph{S2T+TTS}
We transcribe all the S2ST data with open-sourced ASR models (Section~\ref{sec:evaluation}) and train a S2T+TTS system for each language pair. We build 2000 unigram subword units~\citep{kudo2018subword} from the ASR decoded text as the target.
For TTS, we explore two approaches:
\begin{enumerate*}[label=(\arabic*)]
  \item Transformer TTS~\citep{li2019neural}, and
  \item text-to-unit (T2U).
\end{enumerate*}
The Transformer TTS model has a text encoder, a spectrogram decoder and a HiFi-GAN vocoder~\citep{kong2020hifi}. The T2U model is the same model used in preparing reference units for speech normalizer training  (Section~\ref{sec:data}), and we apply the same unit-based vocoder for the S2UT model for unit-to-speech conversion. Both Transformer TTS and T2U are trained with characters as input.

\begin{table*}[ht!]
\centering
\resizebox{\linewidth}{!}{
\begin{tabular}{c|l|ccc|cccc|cccc}
\multicolumn{5}{c|}{} & \multicolumn{4}{c|}{BLEU ($\uparrow$)} & \multicolumn{4}{c}{MOS ($\uparrow$)} \\
\hline
\hline
\multirow{2}{*}{ID} & & tgt & tgt & tgt & \multirow{2}{*}{Es-En} & \multirow{2}{*}{Fr-En} & \multirow{2}{*}{En-Es} & \multirow{2}{*}{En-Fr} & \multirow{2}{*}{Es-En} & \multirow{2}{*}{Fr-En} & \multirow{2}{*}{En-Es} & \multirow{2}{*}{En-Fr} \\
 & & spkemb & SN & text &  &  &  &  &  &  &  &  \\
\hline
1 & S2UT w/ \origunit & \xmark & \xmark & \xmark & 13.1 & 15.4 & 16.4 & 15.8 & 2.32 $\pm$ 0.10 & 2.43 $\pm$ 0.11 & 2.97 $\pm$ 0.14 & 2.41 $\pm$ 0.08 \\
2 & S2UT w/ \origunit & \cmark & \xmark & \xmark & 16.1 & 16.6 & 19.3 & 15.6 & 2.29 $\pm$ 0.11 & 2.25 $\pm$ 0.10 & 3.48 $\pm$ 0.11 & 2.25 $\pm$ 0.06 \\
\hline
3 & S2UT w/ \ftunit & \xmark & 10-min & \xmark & 17.8 & 18.5 & 20.4 & 16.8 & 2.99 $\pm$ 0.07 & 3.16 $\pm$ 0.07 & 3.92 $\pm$ 0.11 & 2.65 $\pm$ 0.08 \\
4 & S2UT w/ \ftunit & \xmark & 1-hr & \xmark & 18.8 & \textbf{20.3} & 21.8 &  \textbf{18.7} & 3.20 $\pm$ 0.09 & 3.26 $\pm$ 0.08 & 4.09 $\pm$ 0.11 & 2.92 $\pm$ 0.09  \\
5 & S2UT w/ \ftunit & \xmark & 10-hr & \xmark & \textbf{18.9} & 19.9 & \textbf{22.7} & \textbf{18.7} & 3.26 $\pm$ 0.08 & 3.27 $\pm$ 0.08 & 4.17 $\pm$ 0.10 & 2.84 $\pm$ 0.08  \\
\hline
6 & S2T + tf TTS & \xmark & \xmark & ASR & 19.2 & 19.8 & 21.7 & 18.5 & 3.23 $\pm$ 0.13 & 3.22 $\pm$ 0.11 & 4.12 $\pm$ 0.11 & 2.44 $\pm$ 0.08  \\
7 & S2T + T2U & \xmark & \xmark & ASR & 19.4 & 19.7 & 21.8 & 18.9 & 3.16 $\pm$ 0.08 & 3.21 $\pm$ 0.07 & 4.11 $\pm$ 0.11 & 2.87 $\pm$ 0.09 \\
\hline
8 & gt + tf TTS & \xmark & \xmark & \xmark & 88.0 & 87.2 & 82.0 & 69.2 & - & - & - & - \\
9 & gt + T2U & \xmark & \xmark & \xmark & 87.9 & 87.1 & 84.6 & 73.8 & - & - & - & - \\
\hline
\end{tabular}}
\caption{\label{tab:s2s_result} BLEU and MOS (reported with 95\% confidence interval) from systems trained in a single run with \vp~S2ST data~\citep{wang-etal-2021-voxpopuli} and evaluated on \epst~\citep{iranzo2020europarl} test sets. The best results from S2UT w/~\ftunit~are highlighted in bold. (tgt spkemb: target speaker embedding, SN: speech normalization, gt: ground truth, tf: Transformer)
}
\end{table*}

\subsection{Evaluation}
\label{sec:evaluation}
To evaluate translation quality, we first use open-sourced ASR models\footnote{En: \url{https://huggingface.co/facebook/wav2vec2-large-960h-lv60-self}, Es: \url{https://huggingface.co/jonatasgrosman/wav2vec2-large-xlsr-53-spanish}, Fr: \url{https://huggingface.co/jonatasgrosman/wav2vec2-large-fr-voxpopuli-french}}
to decode all systems' speech output.
As the ASR output is in lowercase and without digits and punctuation except apostrophes, we normalize the reference text by mapping numbers to spoken forms and removing punctuation before computing BLEU using \textsc{SacreBLEU}~\citep{post2018call}.
To evaluate the naturalness of the speech output, we collect mean opinion scores (MOS) from human listening tests. We randomly sample 200 utterances for each system, and each sample is rated by 5 raters on a scale of 1 (the worst) to 5 (the best).

\subsection{Textless S2ST training}

\paragraph{Speech normalization}
We finetune the mHuBERT model for En, Es and Fr, respectively, resulting in three language-dependent speech normalizers.
We perform CTC finetuning for 25k updates with the Transformer parameters fixed for the first 10k steps.
We use Adam with $\beta_{1}=0.9, \beta_{2}=0.98, \epsilon=10^{-8}$, and 8k warm-up steps and then exponentially decay the learning rate.
We tune the learning rate and masking probabilities on the dev sets based on unit error rate (UER) between the model prediction and the reference target units.

\paragraph{S2UT}
We follow the same model architecture and training procedure in~\citet{lee2021direct}, except having a larger speech encoder and unit decoder with embedding size 512 and 8 attention heads.
We train the models for 600k steps for \vp~S2ST data, and 800k steps for the combination of \vp~and mined data, and use Adam with $\beta_{1}=0.9, \beta_{2}=0.98, \epsilon=10^{-8}$, and inverse square root learning rate decay schedule with 10k warmup steps.
We use label smoothing of 0.2 and tune the learning rate and dropout on the dev set.
The model with the best BLEU on the dev set is used for evaluation.
All S2UT systems including the baselines are trained with an auxiliary task weight of 8.0.

\paragraph{Unit-based vocoder}
We train one vocoder for each language, respectively. All vocoders are trained with~\origunit~sequences as input, since they contain the duration information of natural speech for each unit. We follow the training procedure in~\citet{polyak2021speech} and train for 500k updates with the weight on the MSE loss set to 1.0. 
The vocoder is used for generating speech from either \origunit~or \ftunit, as they originate from the same K-means clustering process.

\section{Results}
\label{sec:exp_result}

\begin{table*}[ht!]
\centering
\resizebox{\linewidth}{!}{
\begin{tabular}{c|l|cccc|cc|cc|c|c}
\multicolumn{6}{c|}{} & \multicolumn{2}{c|}{Es-En} & \multicolumn{2}{c|}{Fr-En} & En-Es & En-Fr \\
\hline
\hline
\multirow{2}{*}{ID} & & \multirow{2}{*}{data} & tgt & tgt & tgt & \multirow{2}{*}{EP} & \multirow{2}{*}{CVST} & \multirow{2}{*}{EP} & \multirow{2}{*}{CVST} & \multirow{2}{*}{EP} & \multirow{2}{*}{EP} \\
 & & & spkemb & SN & text &  &  &  &  &  &  \\
\hline
4 & S2UT w/~\ftunit & VP & \xmark & 1-hr & \xmark & 18.8 & 9.2 & 20.3 & 9.6 & 21.8 & 18.7 \\
\hline
10 & S2UT w/~\origunit & VP+mined & \xmark & \xmark & \xmark & 16.7 & 12.0 & 17.2 & \textbf{16.7} & 19.9 & 18.2  \\
11 & S2UT w/~\origunit & VP+mined & \cmark & \xmark & \xmark & 18.2 & \textbf{16.3} & 19.1 & 16.6 & 21.6 & 18.6 \\
\hline
12 & S2UT w/~\ftunit & VP+mined & \xmark & 1-hr & \xmark & \textbf{21.2} & 15.1 & \textbf{22.1} & 15.9 & \textbf{24.1} & \textbf{20.3} \\
\hline
13 & S2T + tf TTS & VP+mined & \xmark & \xmark & ASR & 21.4 & 14.8 & 22.4 & 16.7  & 24.3 & 20.9 \\
14 & S2T + T2U & VP+mined & \xmark & \xmark & ASR & 21.3 & 14.9 & 22.3 & 16.7 & 24.8 & 21.6  \\
15 & S2T~\citep{wang-etal-2021-voxpopuli} + tf TTS & VP+EP+CVST & \xmark & \xmark & Oracle & 26.0 & 27.3 & 28.1 & 27.7 & - & -  \\
16 & S2T~\citep{wang-etal-2021-voxpopuli} + T2U & VP+EP+CVST & \xmark & \xmark & Oracle & 26.0 & 26.9 & 28.1 & 27.3 & - & - \\
\hline
8 & gt + tf TTS & \xmark & \xmark & \xmark & \xmark & 88.0 & 80.7 & 87.2 & 77.3 & 82.0 & 68.6 \\
9 & gt + T2U & \xmark & \xmark & \xmark & \xmark & 87.9 & 78.8 & 87.1 & 75.9 & 84.6 & 73.8 \\
\hline
\end{tabular}}
\caption{\label{tab:mined_result} BLEU scores ($\uparrow$) from systems trained in a single run with the combination of \vp~S2ST data (VP)~\citep{wang-etal-2021-voxpopuli} and mined S2ST data~\citep{duquenne2021multimodal} and evaluated on \epst~(EP)~\citep{iranzo2020europarl} and CoVoST 2 (CVST)~\citep{wang2020covost} test sets.
The S2T model in~\citet{wang-etal-2021-voxpopuli} is trained on more than 500 hrs of S2T data.
The best results from S2UT with VP+mined data are highlighted in bold. (tgt spkemb: target speaker embedding, SN: speech normalization, gt: ground truth, tf: Transformer)}
\end{table*}

\subsection{Textless S2ST}
\paragraph{S2ST with supervised data}
\autoref{tab:s2s_result} summarizes the results from systems trained with \vp~S2ST data.
We also list the results from applying TTS on the ground truth reference text (8, 9) to demonstrate the impact from ASR errors and potentially low quality speech on the BLEU score.

First, compared with the basic setup, the baseline with target speaker embedding can give a 1.2-3 BLEU improvement on three language pairs (1 vs.~2), implying that there exists variations in \origunit~sequences which are hard to model without extra information from the target speech signals.
However, with only 10 minutes of paired multiple-to-single speaker speech data, we obtain \ftunit~that improves S2UT model performance by 1.5 BLEU on average (2 vs.~3).
The translation quality improves as we increase the amount of parallel data for training the speech normalizer. In the end, with 10 hours of finetuning data, we obtain an average 4.9 BLEU gain from the four language pairs compared to the basic setup (1 vs.~5).

On the other hand, compared with S2T+TTS systems that uses extra ASR models for converting speech to text for training the translation model (6, 7), our best textless S2ST systems (5) can perform similarly to text-based systems without the need of human annotations for building the ASR models.

We see that the MOS of S2UT systems trained with~\origunit~is on average 0.85 lower than that of systems trained with~\ftunit~(1 vs.~5).
We notice that the former often produces stuttering in the output speech, a potential cause to lower MOS.
While worse audio quality may affect ASR-based evaluation and lead to lower BLEU, we verify that this was not the case as the ASR models could still capture the content.
We also see that the proposed textless S2ST system can produce audios with similar naturalness as Transformer TTS models (5 vs.~6).

\paragraph{S2ST with supervised data and mined data}
Next, we add the mined S2ST data for model training, and the results are summarized in~\autoref{tab:mined_result}. We apply the speech normalizer trained with 1-hr data, as it provides similar translation performance as a speech normalizer trained with 10-hr data in \vp-only experiments (4 vs.~5 in~\autoref{tab:s2s_result}).

On the \epst~test set, we see consistent trend across the S2UT models trained with~\ftunit~and the two baselines with~\origunit, where the proposed approach gives on average 3.9 BLEU improvement compared to the basic setup (10 vs.~12), indicating that the speech normalizer trained on \vp~and Common Voice data can also be applied to audios from different domains, e.g.~LibriVox, where the mined data is collected.
The addition of mined data with the proposed speech normalization technique achieves an average of 2.0 BLEU gain over four language directions (4 vs.~12).

We also examine model performance on the CoVoST 2 test set~\citep{wang2020covost} and see even larger improvements brought by mined data (10, 11, 12 vs.~4). One possible reason for this is that LibriVox is more similar to the domain of CoVoST 2 than that of \epst.
With target speaker embedding, mined data improves S2ST by 7.1 BLEU on average (4 vs.~11).
S2UT with~\ftunit~does not perform as well, and one explanation is that we select the best model based on the \epst~dev set during model training.

Compared with S2T+TTS systems trained with text obtained from ASR, there is an average of 0.6 BLEU gap from our proposed system on \epst~test sets (12 vs.~14).
As the En ASR model was trained on Libripeech~\citep{panayotov2015librispeech}, it can decode high quality text output for the mined data.
We also list results from the S2T systems from~\citet{wang-etal-2021-voxpopuli}\footnote{Models downloaded from \url{https://github.com/facebookresearch/voxpopuli/}.} (15, 16), which shows the impact of having oracle text and in-domain training data and serves as an upper bound for the textless S2ST system performance.

\subsection{Analysis on the speech normalizer}
We analyze \ftunit~to understand how the speech normalization process helps improve S2UT performance.
First, to verify that the process preserves the lexical content, we perform a speech resynthesis study as in~\citet{polyak2021speech}.
We use the \vp~ASR test sets, run the unit-based vocoder with different versions of discrete units extracted from the audio as input, and compute word error rate (WER) of the audio output.
In addition to comparing between \ftunit~and \textit{reduced} \origunit, 
we list the WER from the original audio to demonstrate the quality of the ASR models and the gap caused by the unit-based vocoder.

We see from \autoref{tab:sn_wer} that \ftunit~from a speech normalizer finetuned on 1-hr data achieves similar WER as \origunit, indicating that the normalization process does not change the content of the speech.
In addition, we observe that \ftunit~sequences are on average $15\%$ shorter than \textit{reduced} \origunit~sequences. 
We find that this is mainly due to the fact that the speech normalizer does not output units for the long silence in the audio, while \textit{reduced}~\origunit~encodes non-speech segments such as silence and background noises.
Therefore, \ftunit~is a shorter and cleaner target for training S2UT models.

Next, to examine that the speech normalizer reduces variations in speech across speakers, we sample 400 pairs of audios from Common Voice~\citep{ardila2020common} for En, Es and Fr, respectively. Each pair contains two speakers reading the same text prompt.
\autoref{tab:sn_uer} shows the unit error rate (UER) between the unit sequences extracted from the paired audios.
We see that \ftunit~has UER that is on average $58\%$ of the UER of \textit{reduced} \origunit, showing that \ftunit~has less variations across speakers.

\begin{table}[t]
\centering
\begin{tabular}{l|ccc}
WER ($\downarrow$) & En & Es & Fr \\
\hline
original audio  & 14.2 & 15.5 & 18.5 \\
\hline
\textit{reduced} \origunit  & 22.4 & 22.7 & 24.1 \\
\ftunit~(10-min)   & 23.5 & 25.3 & 31.7 \\
\ftunit~(1-hr)   & 21.2 & 20.5 & 24.6 \\
\ftunit~(10-hr)   & 22.0 & 25.3 & 24.2 \\
\end{tabular}
\caption{\label{tab:sn_wer} Speech resynthesis results on the \vp~ASR test set.}
\end{table}

\begin{table}[t]
\centering
\begin{tabular}{l|ccc}
UER ($\downarrow$) & En & Es & Fr \\
\hline
\textit{reduced} \origunit & 74.4 & 70.6 & 73.5 \\
\ftunit~(1-hr) & 48.2 & 31.6 & 46.4 \\
\end{tabular}
\caption{\label{tab:sn_uer} Unit error rate (UER) between units extracted from 400 pairs of audios from the Common Voice dataset. }
\end{table}

\subsection{Analysis of mined data}
\begin{figure}
    \centering
    \begin{tikzpicture}
\begin{axis}[
  xlabel=amount of mined data (hrs),
  ylabel=BLEU,
  ymax=22.0,
  width=1.0\columnwidth,
  height=3.5cm
]
\addplot table [y=bleu, x=hours]{mined_threshold.dat};
\node [above, font=\small] at (axis cs:  441.7, 21.2 ) {t=1.06};
\node [above, font=\small] at (axis cs:  336.5, 20.8 ) {t=1.065};
\node [below, font=\small] at (axis cs:  255.4, 19.9 ) {t=1.07};
\node [above, font=\small] at (axis cs:  192.8, 20.2 ) {t=1.075};
\node [left, font=\small] at (axis cs:  144.0, 20.4 ) {t=1.08};
\end{axis}
\end{tikzpicture}
    \caption{BLEU scores ($\uparrow$) on \epst~Es-En test set from models trained with~\vp~and mined data filtered at different thresholds (t) for the similarity score.}
    \label{fig:mined_threshold}
\end{figure}
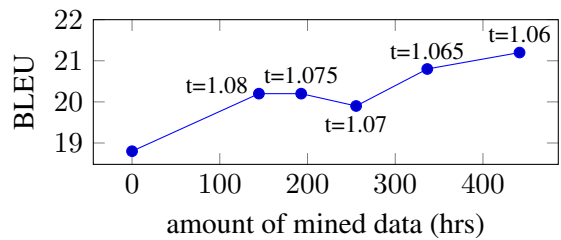
Each pair of aligned speech in the mined data has an associated semantic similarity score. In experiments above, we set the score threshold as $1.06$, and use all mined data with scores above it. Given the trade-off between the quality and quantity of mined data, we analyze how the S2ST performance changes with the threshold set in mined data selection.~\autoref{fig:mined_threshold} demonstrates BLEU scores on \epst~Es-En test set from S2UT systems trained with 1-hr~\ftunit.
The mined data is useful at different thresholds given its gains over the model trained without mined data. As we increase the threshold from $1.06$ to $1.07$, the performance drops due to less training data.

\section{Conclusion}
\label{sec:conclusion}
We present a textless S2ST system that can be trained with real target speech data.
The key to the success is a self-supervised unit-based speech normalization process, which reduces variations in the multi-speaker target speech while retaining the lexical content.
To achieve this, we take advantage of self-supervised discrete representations of a reference speaker speech and perform CTC finetuning with a pre-trained speech encoder.
The speech normalizer can be trained with one hour of parallel speech data without the need of any human annotations and works for speech in different recording conditions and in different languages.
We conduct experiments on the \vp~S2ST dataset and the mined speech data to empirically demonstrate its usefulness in improving S2ST system translation quality for the first time.
In the future, we plan to investigate more textless approaches to improve model performance such as self-supervised pre-training.
All the experiments and ASR evaluation are conducted with public datasets or open-sourced models.

\section*{Acknowledgements}
The authors would like to thank Adam Polyak and Felix Kreuk for initial discussions on accent normalization.

\bibliography{refs}
\bibliographystyle{acl_natbib}

\clearpage
\newpage 

\appendix
\section{mHuBERT Training details}
\autoref{tab:mhubert_training} lists the details for the three iterations of mHuBERT training.

\begin{table}[ht]
\centering
\resizebox{\linewidth}{!}{
\begin{tabular}{c|c|c}
iteration & target features & K-means \\
\hline
\hline
1 & MFCC & 100 \\
2 & 6-th layer from the first iteration & 500 \\
3 & 9-th layer from the second iteration & 500 \\
\end{tabular}}
\caption{\label{tab:mhubert_training} Setup for the target labels used in mHuBERT training.}
\end{table}


\section{Unit-based Vocoder}
\autoref{tab:vocoder_resyn} shows the resynthesis performance of the unit-based vocoder of each language. The WER on the original audio indicates the quality of the open-sourced ASR model we use for evaluation. The WER difference between original audio and \origunit~shows the quality of the vocoder, and the difference between \origunit~and \textit{reduced} \origunit~shows the further impact brought by the duration prediction module.

\begin{table}[ht]
\centering
\begin{tabular}{c|ccc}
WER ($\downarrow$) & En & Es & Fr \\
\hline
\hline
original audio & 2.0 & 8.4 & 24.0 \\
\origunit & 2.8 & 12.0 & 29.3 \\
\textit{reduced} \origunit & 3.4 & 11.9 & 31.3 \\
\end{tabular}
\caption{\label{tab:vocoder_resyn} WER on the TTS dev sets (LJSpeech for En, and CSS10 for Es and Fr) of the audios resynthesized from units.}
\end{table}

\section{Text-to-Unit (T2U)}
\autoref{tab:t2u_wer} lists the WER of the audios generated by the T2U model, which is used in generating the reference target units for speech normalizer training.
As the T2U model is trained with \textit{reduced} unit sequences as the target, during synthesis, we apply the unit-based vocoder with duration prediction.
We can see that T2U with a unit-based vocoder can produce high quality audio and can serve as another option of TTS.

\begin{table}[ht]
\centering
\begin{tabular}{c|ccc}
WER ($\downarrow$) & En & Es & Fr \\
\hline
\hline
original audio & 2.0 & 8.4 & 24.0 \\
T2U & 4.2 & 9.1 & 24.4 \\
\end{tabular}
\caption{\label{tab:t2u_wer} WER on the TTS dev sets (LJSpeech for En, and CSS10 for Es and Fr).}
\end{table}

\section{Hyper-parameters}
\autoref{tab:hp_sn} lists the best hyper-parameters for training the speech normalizers for the three languages and three data setups, respectively. All models are trained on 8 GPUs with a batch size of 100-second (maximum total input audio length).

\autoref{tab:hp_s2ut} lists the best learning rate tuned on the dev set for the S2UT experiments listed in~\autoref{tab:s2s_result} and~\autoref{tab:mined_result}. All models are trained on 8 GPUs with a total batch size of 160k tokens and dropout of 0.3, except for Es-En experiment ID 1 which uses 0.1.

\begin{table}[ht]
\centering
\resizebox{\linewidth}{!}{
\begin{tabular}{cc|ccc}
\multirow{2}{*}{language} & \multirow{2}{*}{duration} & learning & mask & mask channel \\
 & & rate & prob & prob \\
\hline
\hline
En & 10-min & $0.00003$ & $0.75$ & $0.75$ \\
En & 1-hr & $0.00005$ & $0.5$ & $0.5$ \\
En & 10-hr & $0.0001$ & $0.5$ & $0.75$ \\
\hline
Es & 10-min & $0.00003$ & $0.5$ & $0.75$\\
Es & 1-hr & $0.00003$ & $0.5$ & $0.25$ \\
Es & 10-hr & $0.00005$ & $0.5$ & $0.5$ \\
\hline
Fr & 10-min & $0.00003$ & $0.5$ & $0.5$ \\
Fr & 1-hr & $0.00005$ & $0.5$ & $0.25$ \\
Fr & 10-hr & $0.00005$ & $0.5$ & $0.25$ \\
\end{tabular}}
\caption{\label{tab:hp_sn} Hyper-parameters for training the speech normalizers.}
\end{table}

\begin{table}[ht]
\centering
\begin{tabular}{c|cccc}
ID & Es-En & Fr-En & En-Es & En-Fr \\
\hline
\hline
1 & 0.0005 & 0.0003 & 0.0003 & 0.0003  \\
2 & 0.0003 & 0.0003 & 0.0003 & 0.0003  \\
3 & 0.0003 & 0.0003 & 0.0003 & 0.0003 \\
4 & 0.0003 & 0.0003 & 0.0003 & 0.0003 \\
5 & 0.0003 & 0.0003  & 0.0003 & 0.0003 \\
10 & 0.0005 & 0.0005 & 0.0005 & 0.0005 \\
11 & 0.0005 & 0.0003 & 0.0005 & 0.0005 \\
12 & 0.0005 & 0.0005 & 0.0005 & 0.0005 \\
\end{tabular}
\caption{\label{tab:hp_s2ut} Learning rate for S2UT model training.}
\end{table}

\section{Dev BLEU}
\autoref{tab:dev_bleu} shows the BLEU scores on the~\epst~dev sets from systems in~\autoref{tab:s2s_result} and~\autoref{tab:mined_result}. 

\begin{table}[ht]
\centering
\begin{tabular}{c|cccc}
ID & Es-En & Fr-En & En-Es & En-Fr \\
\hline
\hline
1 & 15.4 & 16.0 & 15.9 & 14.7  \\
2 & 18.4 & 17.4 & 19.1 & 15.5 \\
3 & 20.5 & 19.8 & 20.5 & 16.2 \\
4 & 21.4 & 21.0 & 20.8 & 17.6 \\
5 & 21.6 & 21.1 & 22.0 & 17.8 \\
7 & 22.3 & 20.5 & 21.8 & 18.0 \\
10 & 19.0 & 18.7 & 19.8 & 17.2 \\
11 & 20.5 & 20.7 & 20.8 & 17.8 \\
12 & 23.8 & 23.7 & 23.8 & 19.3 \\
14 & 23.7 & 23.6 & 25.0 & 20.6 \\
16 & 28.6 & 29.1 & - & - \\
\end{tabular}
\caption{\label{tab:dev_bleu} BLEU scores on the~\epst~dev sets }
\end{table}

\end{document}